\def\year{2018}\relax
\documentclass[letterpaper]{article}
\usepackage{aaai18}
\usepackage{times}
\usepackage{lipsum}
\usepackage{helvet}
\usepackage{courier}
\usepackage{mathtools}
\usepackage{amsmath}
\usepackage{latexsym}
\usepackage{float}
\usepackage{graphicx}
\usepackage{algorithm}
\usepackage{algpseudocode}
\usepackage{amssymb}
\usepackage{array}
\usepackage{multirow}
\usepackage{booktabs}
\usepackage{siunitx}
\usepackage{url}
\usepackage{tablefootnote}
\usepackage{amsthm}
\usepackage{amssymb}
\usepackage{pifont}
\newcommand{\cmark}{\ding{51}}%
\newcommand{\xmark}{\ding{55}}%
\usepackage{array}
\usepackage{xcolor}
\usepackage{soul}

\DeclareMathOperator*{\argmax}{arg\,max}

\newcommand*\samethanks[1][\value{footnote}]{\footnotemark[#1]}
\definecolor{8red}{HTML}{A70000}
\definecolor{6red}{HTML}{FF0000}
\definecolor{4red}{HTML}{FF5252}
\definecolor{2red}{HTML}{FF7B7B}
\definecolor{1red}{HTML}{FFBABA}
\DeclareRobustCommand{\hleight}[1]{{\sethlcolor{8red}\hl{#1}}}
\DeclareRobustCommand{\hlsix}[1]{{\sethlcolor{6red}\hl{#1}}}
\DeclareRobustCommand{\hlfour}[1]{{\sethlcolor{4red}\hl{#1}}}
\DeclareRobustCommand{\hltwo}[1]{{\sethlcolor{2red}\hl{#1}}}
\DeclareRobustCommand{\hlone}[1]{{\sethlcolor{1red}\hl{#1}}}
\DeclarePairedDelimiter{\abs}{\lvert}{\rvert}
\DeclarePairedDelimiter\norm{\lVert}{\rVert}%

\title{Learning to Attend via Word-Aspect Associative Fusion \\ for Aspect-based Sentiment Analysis}

\author{Yi Tay\thanks{Denotes equal contribution}\textsuperscript{$1$}, Luu Anh Tuan\samethanks \textsuperscript{$2$} \and Siu Cheung Hui\textsuperscript{$3$}\\
\textsuperscript{$1,3$}\:Nanyang Technological University \\ School of Computer Science and Engineering, Singapore \\
\textsuperscript{$2$}\:Institute for Infocomm Research, Singapore \\
}

\date{}

\begin{document}
\maketitle
\begin{abstract}
  Aspect-based sentiment analysis (ABSA) tries to predict the polarity of a given document with respect to a given aspect entity. While neural network architectures have been successful in predicting the overall polarity of sentences, aspect-specific sentiment analysis still remains as an open problem. In this paper, we propose a novel method for integrating aspect information into the neural model. More specifically, we incorporate aspect information into the neural model by modeling word-aspect relationships. Our novel model, \textit{Aspect Fusion LSTM} (AF-LSTM) learns to attend based on associative relationships between sentence words and aspect which allows our model to adaptively focus on the correct words given an aspect term. This ameliorates the flaws of other state-of-the-art models that utilize naive concatenations to model word-aspect similarity. Instead, our model adopts circular convolution and circular correlation to model the similarity between aspect and words and elegantly incorporates this within a differentiable neural attention framework. Finally, our model is end-to-end differentiable and highly related to convolution-correlation (holographic like) memories. Our proposed neural model achieves state-of-the-art performance on benchmark datasets, outperforming ATAE-LSTM by $4\%-5\%$ on average across multiple datasets.
\end{abstract}

\section{Introduction}
Sentiment analysis lives at the heart of many business and social applications which explains its wild popularity in NLP research. Aspect-based sentiment analysis (ABSA) goes deeper by trying to predict polarity with respect to a specific aspect term. For example, consider the following review, \textit{`I love the user interface but this app is practically useless!'}. Clearly, we observe that there are two aspects (user interface and functionality) with completely opposite polarities. As such, techniques that are able to incorporate aspect for making predictions are not only highly desirable but also significantly more realistic compared to coarse-grained sentiment analysis. Recently, end-to-end neural networks (or deep learning) \cite{DBLP:conf/emnlp/WangHZZ16,DBLP:conf/wsdm/LiGM17} such as the long short-term memory networks \cite{hochreiter1997long} and memory networks \cite{DBLP:conf/nips/SukhbaatarSWF15} have demonstrated promising performance on ABSA tasks without requiring any laborious feature engineering. 

The task of ABSA introduces a challenging problem of incorporating aspect information into neural architectures. As such, deep learning architectures that are able to elegantly incorporate aspect information together with sentence modeling are highly desirable. Recently, there have been a myriad of models proposed for this purpose. For example, ATAE-LSTM \cite{DBLP:conf/emnlp/WangHZZ16} is a recently incepted attention based model that learns to attend to different parts of the sentence given the aspect information. ATAE-LSTM tries to incorporate aspect information by adopting a simple concatenation of context words and aspect. This is done both at the attention layer and the sentence modeling layer (inputs to the LSTM). Consequently, the ATAE-LSTM model suffers from the following drawbacks:

\begin{itemize}
\item Instead of allowing the attention layer to focus on learning the relative importance of context words, the attention layer is given the extra burden of modeling the relationship between aspect and context words.
\item The parameters of LSTM are now given an extra burden aside from modeling sequential information, i.e., it has to also learn relationships between aspect and words. The LSTM layer in ATAE-LSTM is being trained on a sequence that is dominated by the aspect embedding. As such, this would make the model significantly harder to train. 
\item Naive concatenation doubles the input to the LSTM layer in ATAE-LSTM which incurs additional parameter costs to the LSTM layer. This has implications in terms of memory footprint, computational complexity and risk of overfitting. 
\end{itemize}

In summary, the important question here is whether the naive concatenation of aspect and words at \textit{both} the LSTM layer and attention layer is necessary or even desirable. In fact, our early empirical experiments showed that the ATAE-LSTM does not always outperform the baseline LSTM model. We believe that this is caused by the word-aspect concatenation making the model difficult to train. As such, this paper aims to tackle the weaknesses of ATAE-LSTM while maintaining the advantages of aspect-aware attentions. Our model cleverly separates the responsibilities of layers by incorporating a dedicated \textit{association layer} for first modeling the relationships between aspect and context words, and then allowing the attention layer to focus on learning the relative importance of the fused context words. As such, the primary goal of this work is to design more effective and efficient attention mechanisms that are aspect-aware.

\subsection{Our Contributions}
The prime contributions of this paper are as follows:
\begin{itemize}
\item We propose a simple and efficient attention mechanism for incorporating aspect information into the neural model for performing aspect-based sentiment analysis. 
\item For the first time, we introduce a novel association layer. In this layer, we adopt \textit{circular convolution of vectors} for performing word-aspect fusion, i.e., learning relationships between aspect and words in a sentence. Our association layer is inspired by the rich history of holographic reduced representation \cite{DBLP:journals/tnn/Plate95} and can be considered as a compressed tensor product. This allows rich higher order relationships between words and aspect to be learned.
\item Overall, we propose \textit{Aspect Fusion LSTM} (AF-LSTM), a novel deep learning architecture, specifically for the task of aspect-based sentiment analysis. Our model achieves not only state-of-the-art performance on benchmark datasets but also significant improvement over many other neural architectures. 
\end{itemize}

\section{Related Work}
Sentiment analysis is a long standing problem in the field of NLP. Simply speaking, this task can be often interpreted as a multi-class (or binary) classification problem in which many decades of research have been dedicated to building features and running them through Support Vector Machine (SVM) classifiers. These traditional features include from sentiment lexicons \cite{rao2009semi,kaji2007building} to ngram features or parse-tree features \cite{DBLP:conf/semeval/KiritchenkoZCM14,DBLP:journals/jair/KiritchenkoZM14}.  

Today, neural architectures are incredibly fashionable for many NLP tasks and clearly, the field of sentiment analysis is of no exception, i.e., the task of document-level sentiment analysis is dominated by neural network architectures \cite{DBLP:journals/corr/BradburyMXS16,tai2015improved,DBLP:conf/acl/QianHLZ17}. The problem our architecture is targeted at is fine-grained sentiment analysis (or aspect-based sentiment analysis) whereby there is an additional complexity in fusing aspects with sentence representations. In order to incorporate aspect information, several architectures have been proposed including the target-dependent LSTM \cite{DBLP:conf/coling/TangQFL16} which models each sentence towards the aspect target. The works most relevant to ours are ATAE-LSTM and AT-LSTM \cite{DBLP:conf/emnlp/WangHZZ16} which are attentional models inspired by \cite{rocktaschel2015reasoning}. AT-LSTM can be considered as a modification of the neural attention of \cite{rocktaschel2015reasoning} for entailment detection that swaps the premise's last hidden state for the aspect embedding. 

Our work is concerned with associative compositional operators which have a rich history in holography. Specifically, in holographic reduced representations and holographic recurrent networks \cite{DBLP:journals/tnn/Plate95,DBLP:conf/nips/Plate92}, circular correlation and convolution are used as encoding-decoding operations which are analogous to storage and retrieval in associative memory models. These associative memory compositional operators can also be interpreted as compressed tensor products that enable second order relationships between word and aspect embeddings to be learned. Moreover, these operators are also efficient with only a computational cost of $\mathcal{O}(n\log n)$ by exploiting computation in the frequency domain, i.e., Fast Fourier Transforms. These associative memory models, though proposed long ago, have seen recent revival in several recent works, e.g., relational learning on knowledge bases \cite{DBLP:conf/aaai/NickelRP16}, question answering \cite{DBLP:conf/sigir/TayPLH17} and even within the recurrent cell  \cite{DBLP:conf/icml/DanihelkaWUKG16}. As such, our work leverages these recent advances and adopts them to ameliorate the weaknesses of ATAE-LSTM by adopting rich second-order associative fusion of aspect and context words. 

It is worthy to mention that a separate class of neural architectures, known as MemNN or End-to-end Memory Network \cite{DBLP:conf/nips/SukhbaatarSWF15}, has also been used for ABSA. Specifically, this frames ABSA as a question answering problem where the network reasons with the aspect as a query and context words as the external memory. \cite{DBLP:conf/emnlp/TangQL16} introduced and applied multi-hop MemNN to ABSA and additionally included a novel mechanism of location attention. On the other hand, \cite{DBLP:conf/wsdm/LiGM17} proposed multi-task MemNN that is also trained on not only polarity prediction but also target detection. A recent work, the Dyadic MemNN \cite{DBLP:conf/cikm/TayTH17} applies rich compositional operators, leveraging neural tensor layers and associative layers on top of memory networks to improve performance on the ABSA task. However, the overall architecture in this paper differs significantly, integrating associative operators into an attention-based LSTM framework instead. 

\section{Our Model}
\begin{figure*}[ht]
  \centering
    \includegraphics[width=0.84\linewidth]{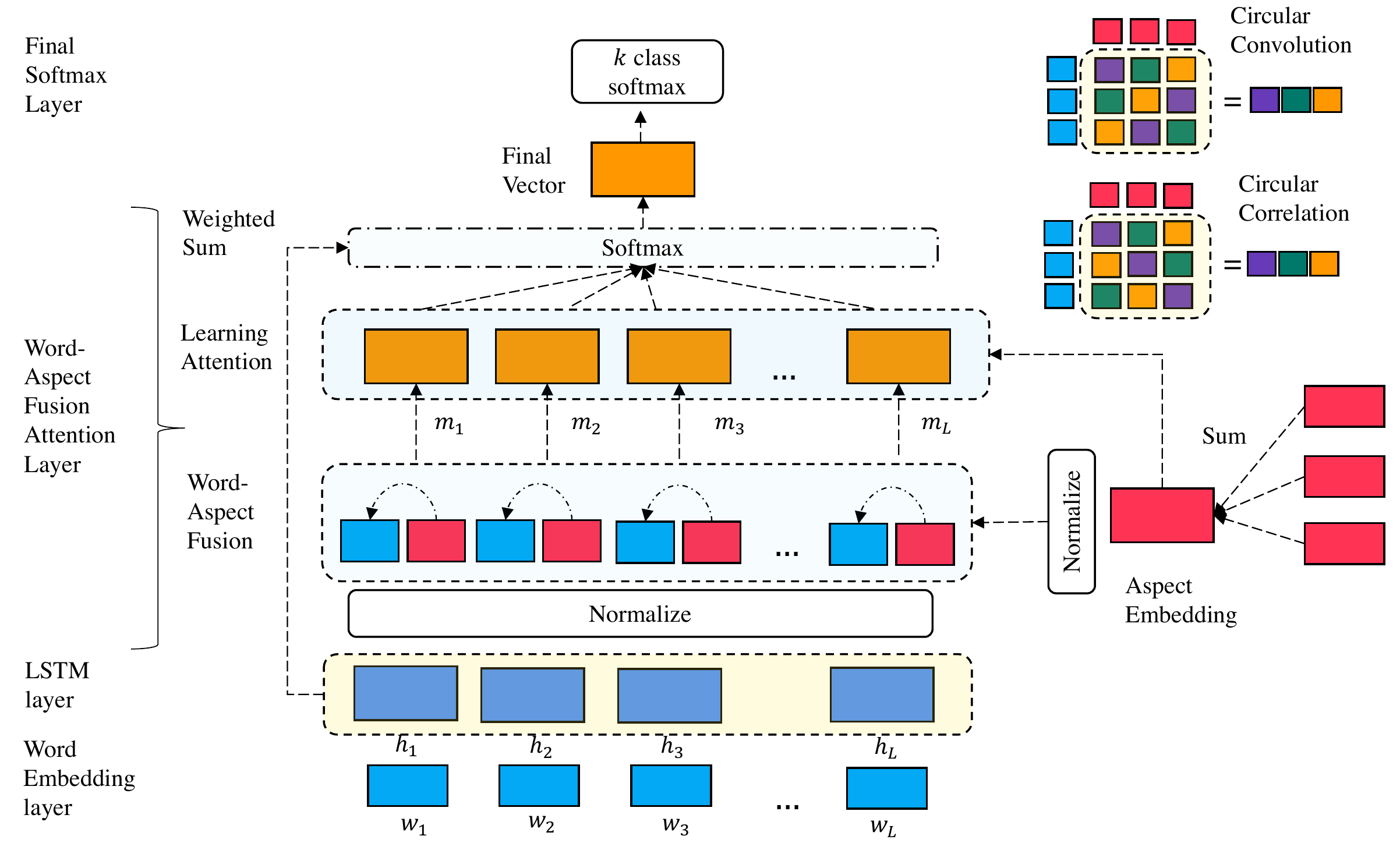}
    \caption{Illustration of our proposed AF-LSTM deep learning architecture (\textit{best viewed in color}). Illustration of circular convolution and circular correlation is depicted in the top right of the image for $d=3$. Compression (summation) patterns are denoted by the matching colors. }
    \label{fig:4B}
\end{figure*}

In this section, we describe our deep learning architecture layer-by-layer. The overall model architecture is illustrated in Figure \ref{fig:4B}.

\subsection{Word Embedding Layer}
The input to our model is a sentence (sequence of words) along with an aspect word or phrase which are passed in as integer values and indexed into the embedding matrix. The word embedding layer is simply a $W_E \in \mathbb{R}^{k \times v}$ where $k$ is the dimension of the word embeddings and $v$ is the vocabulary size. As such, each input sentence is converted into a sequence of $k$ dimensional vectors by the embedding layer. For aspect terms with more than one word, we simply apply a neural bag of words model (sum operator) to learn a single $k$ dimensional vector of the aspect. We refer this as $s$ in this paper.

\subsection{Long Short-Term Memory (LSTM) Layer}
The word representations are then fed into a long short-term memory (LSTM) network \cite{hochreiter1997long}. The operations of the LSTM cell can be described as follows:
\begin{equation}
h_{t} = LSTM(h_{t-1}, x_i)
\end{equation}
where $x_t$ and $h_{t-1}$ are the input vectors at time $t$. The LSTM model is parameterized by output, input and forget gates, controlling the information flow within the recursive operation. For the sake of brevity, we omit the technical details of LSTM which can be found in many related works. 
The output of this layer is a sequence of hidden vectors $\textbf{H} \in \mathbb{R}^{L \times d}$ where $L$ is the maximum sequence length and $d$ is the dimensional size of LSTM. 

\subsection{Word-Aspect Fusion Attention Layer}
In this section, we describe our novel word-aspect fusion attention layer. First, we provide an overview of the mechanics of this new neural building block. 

\begin{itemize}

\item The inputs to this layer are the outputs of the LSTM layer $\textbf{H} \in \mathbb{R}^{d \times L}$ and the aspect embedding $s$.
\item For each output vector $h_i \in \textbf{H}$ from the LSTM layer, we learn a joint representation of word and aspect embedding. This embedding is referred to as $m_i \in \mathbb{R}^{d}$ (memory traces) which encodes the relationship between the context word $h_i$ and aspect. In order to learn a joint representation of $h$ and $s$, we employ associative memory operators to learn $m_i$. 
\item Subsequently, the sequence of encoded joint representations $m_1, m_2 \dots m_L$ is then passed into the attention layer. 
\item The attention layer learns to attend based on $\textbf{M} \in \mathbb{R}^{d \times L}$ instead of $\textbf{H} \in \mathbb{R}^{d \times L}$.

\end{itemize}
In this section, we describe the key steps that are crucial to the operation of our proposed attention mechanism. 

\subsubsection{Associative Memory Operators}

 We first describe the key concepts of associative memory operators which we use to model the relationship between context words and aspect embedding. Let $h_t$ be the hidden state of the LSTM at time $t$ and $s$ be the aspect embedding. In our case, we define two associative compositional operators, namely \textit{circular correlation} (denoted $\star$) and \textit{circular convolution} (denoted $\ast$), as our associative memory operators. The first operator, circular correlation is defined as follows:
\begin{equation}
[h \star s]_k = \sum_{i=0}^{d-1} h_i\:s_{(k+i) \: \bmod\:d}
\end{equation}
where $\star :  \mathbb{R}^{d} \times \mathbb{R}^{d} \rightarrow \mathbb{R}^{d}$ denotes the circular correlation operator. For notational convenience, the subscript $t$ for $h$ is omitted. We also use zero-indexed vectors. Circular correlation can be computed as follows:
\begin{equation}
h \star a = \mathcal{F}^{-1} (\overline{\mathcal{F}(h)} \odot \mathcal{F}(a))
\end{equation}
where $\mathcal{F(.)}$ and $\mathcal{F}^{-1}(.)$ are the Fast Fourier Transform (FFT) and inverse Fast Fourier Transform. $\overline{\mathcal{F}(h)}$ denotes the complex conjugate of $\mathcal{F}(h)$. $\odot$ is the element-wise (or Hadamard) product. Next, we define circular convolution as follows:
\begin{equation}
[h \ast s]_k = \sum_{i=0}^{d-1} h_i\:s_{(k-i) \bmod\:d}
\end{equation}   
where $\ast :  \mathbb{R}^{d} \times \mathbb{R}^{d} \rightarrow \mathbb{R}^{d}$ denotes the circular convolution operator. Similar to circular correlation, circular convolution can be computed efficiently via $\mathcal{F}^{-1} (\mathcal{F}(h) \odot \mathcal{F}(a))$. Note the absence of the conjugate operator. In our model, \textbf{either} circular correlation or convolution may be employed. In our experimental evaluation, we evaluate both variants of our model. However, note that the key difference between these two operators are the commutative property, i.e., correlation is non-commutative while convolution is not. Overall, the outputs of the association layer are defined as follows:

\begin{equation}
\textbf{M}=[[h_1 \circ s], [h_2 \circ s], [h_3 \circ s]...[h_{L} \circ s]]
\end{equation}
 \noindent where $\circ$ is the associative memory operator adopted, $h_i \in \mathbb{R}^{d}$ are the outputs from the LSTM layer and $s$ is the aspect embedding. As such, the dimension of the output association layer is identical to the inputs of the association layer. Note that the association layer is essentially parameterless and does not increase the parameters of the network. 

\subsubsection{Normalization Layer} 
Before learning association of $h_i$ and $s$, we also include an optional normalization layer. Specifically, we normalize\footnote{This is mentioned as a requirement in \cite{DBLP:conf/nips/Plate92,DBLP:journals/tnn/Plate95} but we found this condition is relatively unimportant for our model. An intuitive explanation might be due to the fact that we have already clipped the norm of the gradients to $\leq 1$.} $\abs{h_i}$ and $\abs{s}$ to $\leq 1$.  Alternatively, we could also consider a batch normalization layer \cite{DBLP:conf/icml/IoffeS15} which we found to improve the performance on certain datasets. Overall, we consider the presence of this normalization layer as a hyperparameter to be tuned.

\subsubsection{Learning Attentive Representations}
Next, the composition of each hidden state $h_t$ and the aspect vector $s$ via associative operators are used for learning attentions. 
\begin{align*}
\textbf{Y} &= tanh(\textbf{W}_{y}\: \textbf{M}) &\\
a &= softmax(w^{T}\:\textbf{Y}) &\\
r &= \textbf{H}\:a^{T} &
\end{align*}
\noindent where $\textbf{W}_y \in \mathbb{R}^{d \times d}$ and $w \in \mathbb{R}^{d}$ are the parameters of the attention layer.  $\textbf{H} \in \mathbb{R}^{L \times d}$ is a matrix of the LSTM output, $L$ is the sequence length and $d$ is the dimensionality of the LSTM. $a \in \mathbb{R}^{L}$ is the attention vector.  The attention vector contains a probabilistic weighting of all the hidden representations from the LSTM layer and produces a final weighted representation $r \in \mathbb{R}^{d}$.  Following \cite{DBLP:conf/emnlp/WangHZZ16}, we may also add a projection\footnote{In practice, we found that this layer does not seem to influence the performance much. However, it does not degrade the performance at the very least.} layer that combines the attentive representation with the last hidden state. The final representation is described as follows:
\begin{equation}
r = tanh(W_p \: r + W_x \: h_L)
\end{equation}
where $W_p \in \mathbb{R}^{d \times d}$ and $W_x \in \mathbb{R}^{d \times d}$ are the parameters of this layer. 

\subsection{Final Softmax Layer}
The weighted representation $r$ of the sentence is then passed into the final layer which converts the weighted representation $r$ into a probability distribution. This layer also consists of a linear transformation layer and then a softmax function. The final layer can be defined as follows:
\begin{align*}
x &= W_f \ldotp r + b_f &\\
p(y=k|x) &= \frac{e^{x^T\theta_k}}{\sum^{K}_{k=1}e^{x^T\theta_k}} &
\end{align*}
where $\theta_k$ is the weight vector of the $k$th class. $W_f$ and $b_f$ are the parameters of the final layer. 
\subsection{Optimization and Training}
For optimization, we adopt the cross entropy loss function. 
\begin{equation}
L = -\sum^{N}_{i=1} \: [ y_i \log o_i + (1-y_i)\log(1-o_i)] +R 
\end{equation}
where $o$ is the output of the softmax layer and $R= \lambda\norm{\psi}^{2}_2$. $\psi$ contains all the parameters of the network and $ \lambda\norm{\psi}^{2}_2$ is the L2 regularization. 

\section{Discussion and Analysis}
\label{sec:discussion}
In this section, we provide some intuitions and key advantages behind the usage of associative memory operators in our word-aspect fusion attention mechanism. 
\subsection{Connections to Holographic Memory Models}

Our model is highly related to associative memory models and holographic reduced representations \cite{DBLP:journals/tnn/Plate95} that adopt a series of convolutions and correlations to store and retrieve item pairs. In these models, an encoding operation (e.g., circular convolution) is used to store the association of two vectors:
\begin{equation}
m = h \ast s
\end{equation}
and subsequently a decoding operation is used to retrieve $s$ via:
\begin{equation}
s' \approx h \star m = s \ast (h \star h)
\end{equation}
where $h \star h \approx \delta$ is the identity element of convolution. Given the noisy vector, $s'$, we are able to perform clean up which returns the most similar item in the stored memory $m$:
\begin{align*}
s = \argmax_{s_i} \: s_i^{\intercal} (h \ast m)
\end{align*} 
In our proposed word-aspect fusion attention mechanism, recall that we are learning attentions by using the memory trace $m$ which is formed from composing the aspect embedding with the context, i.e., LSTM outputs $h_i$. In this case, either the circular convolution or circular correlation may be used as the encoding operation. Consider the case where circular convolution is used as the encoding operation, the gradients of the aspects at the attention layer are as follows:
\begin{align*}
\frac{\partial E}{\partial s_i} = \sum_{k} \frac{\partial E}{\partial a_{j}} \: h_{(k-j\bmod d)}
\label{eqn:grad}
\end{align*}
where $a$ is the attention vector and $\sum_{k} \frac{\partial E}{\partial a_{j}} \: h_{(k-j\bmod d)}$ is essentially the circular correlation operation \cite{DBLP:conf/nips/Plate92,DBLP:journals/tnn/Plate95}. As such, when circular convolution is used as the associative memory operator, its inverse (circular correlation) represents the decoding operation. This works vice versa as well, i.e., if circular correlation is the encoding operation, then circular convolution will become the decoding operation. For an explanation to why circular convolution decodes circular correlation, we refer interested readers to \cite{DBLP:journals/tnn/Plate95} for an extensive technical overview. 

Finally, considering the context of an end-to-end neural architecture, the forward operation is the encoding operation and the parameters of the attention layer are updated via the decoding operation which simulates storage and retrieval in holographic convolution-correlation memories, i.e., the process of encoding and decoding is reflected in the forward propagation and back propagation of the network. Intuitively, during the forward propagation (test time), this can be interpreted as trying to retrieve the context word $h_i$ that is closest to the aspect embedding $s$ which forms the crux of our word-aspect fusion attention layer. 

\subsubsection{Capturing Second Order Interactions}
Additionally, we observe several advantages of our model over \cite{DBLP:conf/emnlp/WangHZZ16}. First, our model enables richer representational learning between context words and aspect. Associative memory operators are also known to be compressed tensor products which learn rich relationships between two vectors \cite{DBLP:conf/aaai/NickelRP16}. Figure \ref{fig:4B} (\textit{top right}) shows the visual interpretation of circular convolution and correlation with case $d=3$. Compression over a tensor product is shown by matching colours. For example, the summation of all purple blocks in circular convolution forms the first element of the composed vector $m$.

\subsubsection{On efficiency}
Unlike a simple concatenation employed in \cite{DBLP:conf/emnlp/WangHZZ16}, our model does not increase the parameters of the network and our association layer is actually parameterless. Each operation in the association layer can be computed efficiently with $\mathcal{O}(n\log n)$ which eliminates concerns about scalability. This also does not increase the burden of the LSTM layer or attention layer to learn the relationships between aspects and words. 
\subsubsection{Separation of Goals}
Overall, our model allows a clear separation of goals, i.e., the associations between aspect and words are first learned independently and not coupled with other layers (e.g., LSTM and attention layers) which have their own primary objective such as learning semantic compositionality. 

\begin{table}[htbp]
  \centering
\small
    \begin{tabular}{c|c|cccc}
    \toprule
    \midrule
    
    Task & \textbf{Dataset} & \textbf{\# All} & \textbf{\# Pos} & \textbf{\# Neg} & \textbf{\# Neu} \\
    \midrule
   T& Laptops Train & 1813  & 767   & 673   & 373 \\
    T & Laptops Dev & 499   & 220   & 192   & 87 \\
     T& Laptops Test & 638   & 341   & 128   & 169 \\
    \midrule
     T& Restaurants Train & 3102  & 685   & 1886  & 531 \\
     T& Restaurants Dev & 500   & 278   & 120   & 102 \\
     T& Restaurants Test & 1120  & 728   & 196   & 196 \\
    \midrule
     C& Restaurants Train & 3018  & 1855  & 733   & 430 \\
     C& Restaurants Dev & 500   & 324   & 106   & 70 \\
     C& Restaurants Test & 973   & 657   & 222   & 94 \\
    \midrule
     C& SE 14+15 Train & 3587  & 1069  & 2310  & 208 \\
     C& SE 14+15 Dev & 1011   & 455   & 496   & 60 \\
     C& SE 14+15 Test & 427   & 274  & 134   & 19 \\
    \midrule
    \toprule
    \end{tabular}%
    \caption{Dataset statistics of all datasets. T denotes the term classification task and C denotes the category classification task. }
  \label{tab:dataset}%
\end{table}%

\section{Experimental Evaluation}
\label{sec:exp}
In this section, we introduce our empirical evaluation and comparisons. We introduce the datasets used, experimental protocol and finally the experimental results. We conduct on two tasks, namely aspect term classification and aspect category classification. The main difference is that \textit{term} can be more than one word that is found within the sentence itself. On the other hand, \textit{category} refers to generic labels such as `service' or `food' which may or may not be found in the sentence. Since we treat aspect embeddings to be the sum of all words (a single vector), the experiments can be conducted in the same fashion irregardless of the nature of term or category.

\begin{table*}[htbp]
  \centering

    \begin{tabular}{c|c|cccc|cccc|c}
    \toprule
    \midrule
    & & \multicolumn{4}{c|}{Term Classification} & \multicolumn{4}{c|}{Category Classification} & \\
    \midrule
          &       & \multicolumn{2}{c}{Laptops} & \multicolumn{2}{c|}{Restaurants} & \multicolumn{2}{c}{Restaurants} & \multicolumn{2}{c|}{SemEval 14+15} & \\
          \midrule
    \textbf{Model} & Aspect & 3-way & Binary & 3-way & Binary & 3-way & Binary & 3-way & Binary & Avg\\
    \midrule
    Majority & No    & 53.45 & 72.71 & 65.00 & 78.79 & 67.52 & 74.40 & 64.16 & 75.12 & 68.89\\
    NBOW  & No    & 58.62 & 73.34 & 67.49 & 82.47 & 70.81 & 78.61 & 70.92 & 77.18 & 72.43\\
    LSTM  & No    & 61.75 & 78.25 & 67.94 & 82.03 & 73.38 & 79.97 & 75.96 & 79.92 & 74.90 \\
    TD-LSTM & Yes   & 62.38 & 79.31 & 69.73 & 84.41 & 79.97 & 75.96 & 79.92 & 74.90 & 75.63\\
    AT-LSTM  & Yes   & 65.83 & 78.25 & 74.37 & 84.74 & 77.90 & 84.87 & 76.16 & 81.28 & 77.93\\
    ATAE-LSTM & Yes   & 60.34 & 74.20 & 70.71 & 84.52 & 77.80 & 83.85 & 74.08 & 78.96 & 75.56\\
    \midrule
    AF-LSTM (MUL) & Yes   & 66.14 & 83.37 & 75.35 & 86.47 & 79.96 & 81.71 & 77.44 & 80.44 & 78.86\\
    AF-LSTM (CORR) & Yes   & 64.89 & 79.96 & 74.76 & 86.91 & 80.47 & 86.58 & 74.68 & \textbf{81.60} & 78.73\\
    AF-LSTM (CONV)  & Yes   & \textbf{68.81} & \textbf{83.58} & \textbf{75.44} & \textbf{87.78} & \textbf{81.29} & \textbf{87.26} & \textbf{78.44} & 81.49 & \textbf{80.51}\\
    \midrule
    \toprule
    \end{tabular}%

     \caption{Comparisons of all deep learning architectures on all datasets. Avg column reports macro-averaged results across all datasets. Best performance is in bold face. Our proposed AF-LSTM (CONV) achieves state-of-the-art performance across all four datasets and settings.}
  \label{tab:main_results}%
\end{table*}%

\subsection{Datasets}
For our experimental evaluation, we adopt several subsets from the popular SemEval 2014 \cite{DBLP:conf/semeval/PontikiGPPAM14} task 4 and SemEval 2015 task 12 which are widely adopted benchmarks in many works \cite{DBLP:conf/emnlp/TangQL16,DBLP:conf/emnlp/WangHZZ16}.  We evaluate on two datasets per task which are reported in Table \ref{tab:dataset}. For term classification (T), we use the Laptops and Restaurants datasets from SemEval 2014. For category classification (C), we use the Restaurants dataset from SemEval 2014 and a combined dataset from both SemEval2014 and SemEval2015. The split is obtained from \cite{DBLP:conf/wsdm/LiGM17} and is denoted as \textit{SemEval 14+15} in our experiments.

\subsubsection{Evaluation Protocol}
Apart from aspect term classification and aspect category classification, we conduct our experiments on two settings, namely three-way classification and binary classification (positive and negative). The metric reported is simply the accuracy score following \cite{DBLP:conf/emnlp/WangHZZ16}. Since there is no official development set for the SemEval 2014 task 4, we construct a development\footnote{Our splits can be found at \url{https://github.com/vanzytay/ABSA_DevSplits}.} set from the training set. Specifically, we take $500$ training instances as our development set for tuning model hyperparameters. Unfortunately, many published works \cite{DBLP:conf/emnlp/WangHZZ16,DBLP:conf/emnlp/TangQL16} do not mention the usage of a development set in their evaluation procedure. The usage of a fixed development set in our experiments (for all models) is motivated by the fluctuating performance on the test set per iteration. It is also best practice in evaluating models which is unfortunate that most prior works do not adopt. As such, we encourage future research to use the datasets with development splits. Additionally, we also reimplement all models on the same environment in the spirit of fair comparison.

\subsection{Compared Models}
In this section, we discuss the compared models. 
\subsubsection{Our Models}
There are two variations to AF-LSTM, namely AF-LSTM (CCOR) and AF-LSTM (CONV), depending on the choice of encoding operation. CCOR and CONV denotes the circular correlation and convolution operations respectively. Additionally, we include a new baseline AF-LSTM (MUL) which swaps the encoding and decoding operator with the Hadamard product (elementwise multiplication). This is mainly to observe the effect of our holographic-inspired attentions.

\subsubsection{Baselines}
Additionally, we include other neural baseline models which are used for comparison. The baselines are listed as follows: 
\begin{itemize}
\item \textbf{Neural Bag-of-Words} (NBOW) is simply the sum of all word embeddings in the context. 
\item \textbf{LSTM} (Long Short-Term Memory) is a widely adopted neural baseline for many NLP tasks. In this model, aspect information is not used. 
\item \textbf{TD-LSTM} (Target-Dependent LSTM) considers the aspect by adopting two LSTMs towards the aspect target and passing both outputs through a linear transformation layer. Note that TD-LSTM reverts to the baseline LSTM for \textit{category} classification tasks.
\item \textbf{AT-LSTM} (Attention LSTM) \cite{DBLP:conf/emnlp/WangHZZ16} adopts the attention mechanism in LSTM to produce a weighted representation of a sentence. The aspect embedding is added via linear projection and then concatenation at the attention layer. 
\item \textbf{ATAE-LSTM} (Attention LSTM with Aspect Embedding) \cite{DBLP:conf/emnlp/WangHZZ16} can be considered as an extension of AT-LSTM. In this model, the aspect embedding is concatenated with the input before passing into the LSTM layer.

\end{itemize}
It is good to note that, many of these models, such as the vanilla LSTM and AT-LSTM, serve as good baselines for ablation studies. Additionally, we include a simple Majority baseline that predicts the majority class. For example, if positive instances are the majority class in the training set, the Majority baseline will classify all test instances as positive.

\begin{table*}[ht]
  \centering
\small
    \begin{tabular}{l|c|p{7.5cm}|c}
    \toprule
    \midrule
    \textbf{Model} & \textbf{Aspect} & \textbf{Text} & \multicolumn{1}{l}{\textbf{Correct}} \\
    \midrule
    AT-LSTM (w/o aspect) & Appetizer & The appetizers are \hltwo{okay} but the service is \hlsix{slow}. & \xmark  \\
    AT-LSTM (\citeauthor{DBLP:conf/emnlp/WangHZZ16}) & Appetizer & The appetizers are \hltwo{okay} but the service is \hleight{slow}. & \xmark  \\
    ATAE-LSTM  (\citeauthor{DBLP:conf/emnlp/WangHZZ16}) & Appetizer & The appetizers are \hlfour{okay} but the service is \hlsix{slow}. & \xmark  \\
    AF-LSTM (CONV) & Appetizer & The \hlsix{appetizers} \hltwo{are} \hlone{okay} but the service is slow. &  \cmark \\
    \midrule
    AT-LSTM (w/o aspect) & Service & The \hltwo{appetizers} are \hlfour{okay} \hlone{but} the service is slow. & \xmark  \\
        AT-LSTM (\citeauthor{DBLP:conf/emnlp/WangHZZ16})  & Service & The appetizers are okay but the service is \hlsix{slow}. & \cmark  \\
    ATAE-LSTM (\citeauthor{DBLP:conf/emnlp/WangHZZ16})  & Service & The appetizers are okay but the service is \hleight{slow}. & \cmark  \\
    AF-LSTM (CONV)& Service & The appetizers are okay but the service is \hlsix{slow}. & \cmark \\  
    \midrule
    \toprule
    \end{tabular}%
    \caption{Example case study: inspecting attentions on a particularly difficult contrasting polarity test example on the Restaurants dataset. The intensity of the (red) color denotes the strength of the attention weights.}
  \label{tab:example1}%
\end{table*}%

\subsection{Implementation Details}
We implemented all models in TensorFlow. All models are optimized using the Adam optimizer \cite{DBLP:journals/corr/KingmaB14}  with a learning rate of $10^{-3}$. The regularization factor $\lambda$ is set to $4 \times 10^{-6}$ and the batch size tuned amongst $\{25,50\}$. We apply a dropout of $p=0.5$ after the LSTM (or representation) layer. All models are initialized with $300$ dimension Glove Embeddings (840B tokens) \cite{DBLP:conf/emnlp/PenningtonSM14}. All models are trained for $50$ epochs and the result reported is the test score of the model that performed the best on the development set. We also employ early stopping, i.e., we stop training when performance on the development set does not improve after $10$ epochs. As for preprocessing, we lowercased, filtered non-alphanumeric characters and applied NLTK's word tokenizer. All models are evaluated on a single NVIDIA GTX1060 GPU on a Linux machine.

\subsection{Experimental Results}

Table \ref{tab:main_results} shows our experimental results on ABSA. Firstly, we observe that our proposed AF-LSTM (CONV) outperforms all other neural architectures. In fact, AF-LSTM (CONV) outperforms\footnote{We carefully reimplemented ATAE-LSTM and AT-LSTM but we could not reproduce the good results of \cite{DBLP:conf/emnlp/WangHZZ16}. Results also differ because of the introduction of a development set.} ATAE-LSTM by $3\%-8\%$ on 3-way classification and about $2\%-3\%$ on binary classification. The performance of all AF-LSTM models is generally much higher than ATAE-LSTM. Additionally, we observe that ATAE-LSTM is outperformed by AT-LSTM across all settings. This shows that concatenation of aspect and word before the LSTM layer may significantly degrade performance. Moreover, we found that the performance of ATAE-LSTM may not outperform a baseline LSTM on certain datasets (Laptops). The overall performance of ATAE-LSTM is approximately the same as the baseline LSTM.

 Based on the empirical results of AT-LSTM, the concatenation of aspect and word at the attention layer is sound and shows reasonable improvements over the baseline LSTM. However, a simple Hadamard product (MUL) of word and aspect embedding already outperforms the AT-LSTM. AF-LSTM (MUL) outperforms AT-LSTM marginally ($\approx 1\%$) but significantly outperforms ATAE-LSTM ($\approx 3\%$). Finally, AF-LSTM (CONV) shows improvement over the Hadamard product, i.e., the AF-LSTM (MUL) baseline. On the other hand, the performance of AF-LSTM (CCOR) is similar to that of AF-LSTM (MUL). As such, we also observe that circular convolution is a significantly more effective associative operator as compared to circular correlation. We believe that this might be due to the circular correlation being an asymmetric operator. Words can appear in either aspect or context. Therefore, an explicit modeling of asymmetry may degrade the performance. Overall, the best performance is achieved by word-aspect fusion exploiting circular convolution (CONV).

\subsection{Memory / Parameter Size Analysis}
Finally, the parameter size of AF-LSTM is $\approx 810K$ while AT-LSTM and ATAE-LSTM are $\approx 1.1M$ and $\approx 1.4M$ respectively. Note that all variants of AF-LSTM has the same parameter size. The baseline LSTM has $\approx 720K$ parameters. As such, our proposed AF-LSTM has a smaller parameter size while outperforming AT-LSTM and ATAE-LSTM.

\section{Qualitative Analysis}
We selected a double-polarity example from the Restaurants dataset and visualize the reaction of attention weights with respect to the aspect embedding. We extract the attention vector for AF-LSTM, AT-LSTM and ATAE-LSTM. Additionally, we include a variation of AT-LSTM (w/o aspect) by removing the aspect embedding in order to clearly observe the effect of aspect-aware attentions. The hyperparameter settings remain identical. 

Table \ref{tab:example1} shows the visualized attention weights of the following contexts with respect to a particular aspect. In the first example where the aspect is the Appetizer, we see that the attention of AT-LSTM (w/o aspect), AT-LSTM and ATAE-LSTM are focusing at the wrong words, i.e., the focused words `okay` and `slow' are almost equal. On the other hand, AF-LSTM is able to focus on the correct words - `appetizers are okay' and completely ignore `slow'. We believe that this is due to associations learned between the words `appetizers' and `slow' (appetizers cannot be slow). 

In the second example, the aspect is swapped to Service. This time we see all models that have the aspect information focusing on the right words. AT-LSTM (w/o aspect) is randomly guessing whether to focus on `okay' or `slow' since it does not have any aspect information. Overall, we note that AT-LSTM and ATAE-LSTM of \cite{DBLP:conf/emnlp/WangHZZ16} are able to focus on the right words for the aspect of Service. However, both AT-LSTM and ATAE-LSTM are unable to focus on the right words for \textbf{both} aspects (Service and Appetizer) which is a weakness of the approach, i.e., the concatenation operator makes it difficult to model relationships between aspect and context words. Our approach, however, effectively and correctly switches the focus words when given a different aspect.

\section{Conclusion}
We proposed a novel method for incorporating aspect information for learning attentions. We introduce a novel word-aspect fusion attention layer to first learn associative relationships between aspect and context words for learning attentive representations. We show that learning attentions from associative memory traces of word and aspects are effective. Overall, circular convolution remains highly effective for aspect-word fusion. On the other hand, simple elementwise multiplications remains a strong baseline, outperforming simple concatenation models such as AT-LSTM and ATAE-LSTM. Our model shows an significant improvement in performance compared to multiple strong neural baselines. 
\section{Acknowledgements}
The authors would like to thank anonymous reviewers for their feedback and comments.

\bibliography{references} 
\bibliographystyle{aaai}
\end{document}